\documentclass[10pt,twocolumn,letterpaper]{article}

\usepackage{cvpr}              
\usepackage{svg}
\usepackage{multirow}
\usepackage{booktabs}
\usepackage{tabularray}
\usepackage{tabularx}
\usepackage{graphicx}
\usepackage{colortbl}
\definecolor{cvprblue}{rgb}{0.21,0.49,0.74}
\usepackage[pagebackref,breaklinks,colorlinks,citecolor=cvprblue]{hyperref}

\def\paperID{10688} 
\def\confName{CVPR}
\def\confYear{2024}

\title{Manipulating the Label Space for In-Context Classification}

\author{Haokun Chen, Xu Yang, Yuhang Huang, Zihan Wu, Jing Wang, Xin Geng\\
Key Laboratory of New Generation Artificial Intelligence Technology \& \\ Its Interdisciplinary Applications, (Southeast University), Ministry of Education\\
{\tt\small chenhaokun@seu.edu.cn, xuyang\_palm@seu.edu.cn, HuangYH723@outlook.com}\\
{\tt\small zihanwu\_seu@outlook.com, wangjing91@seu.edu.cn, xgeng@seu.edu.cn}
}

\begin{document}
\maketitle
\begin{abstract}
After pre-training by generating the next word conditional on previous words, the Language Model (LM) acquires the ability of In-Context Learning (ICL) that can learn a new task conditional on the context of the given in-context examples (ICEs). Similarly, visually-conditioned Language Modelling is also used to train Vision-Language Models (VLMs) with ICL ability. However, such VLMs typically exhibit weaker classification abilities compared to contrastive learning-based models like CLIP, since the Language Modelling objective does not directly contrast whether an object is paired with a text. To improve the ICL of classification, using more ICEs to provide more knowledge is a straightforward way. However, this may largely increase the selection time, and more importantly, the inclusion of additional in-context images tends to extend the length of the in-context sequence beyond the processing capacity of a VLM. To alleviate these limitations, we propose to manipulate the label space of each ICE to increase its knowledge density, allowing for fewer ICEs to convey as much information as a larger set would. Specifically, we propose two strategies which are Label Distribution Enhancement and Visual Descriptions Enhancement to improve In-context classification performance on diverse datasets, including the classic ImageNet and more fine-grained datasets like CUB-200. Specifically, using our approach on ImageNet, we increase accuracy from 74.70\% in a 4-shot setting to 76.21\% with just 2 shots. surpassing CLIP by 0.67\%. On CUB-200, our method raises 1-shot accuracy from 48.86\% to 69.05\%, 12.15\% higher than CLIP.
The code is given in \url{https://anonymous.4open.science/r/MLS_ICC}.
\end{abstract}

\begin{figure}[t]
  \centering
  \begin{subfigure}{\linewidth}
    \centering
    \includegraphics[width=\linewidth]{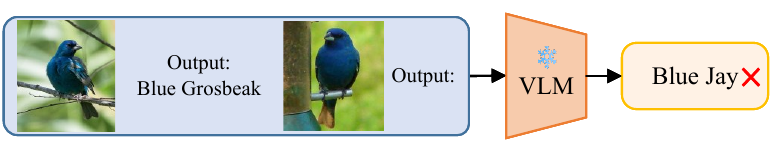}
    \caption{In-context Learning by Single Label.}
    \label{fig:intro-a}
  \end{subfigure}
  \begin{subfigure}{\linewidth}
    \centering
    \includegraphics[width=\linewidth]{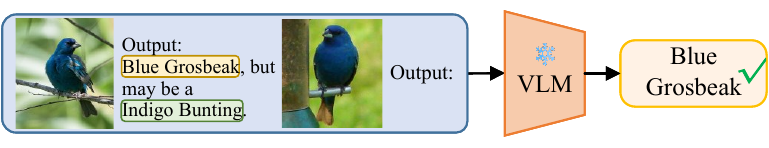}
    \caption{In-context Learning by LDE.}
    \label{fig:intro-b}
  \end{subfigure}
  \begin{subfigure}{\linewidth}
    \centering
    \includegraphics[width=\linewidth]{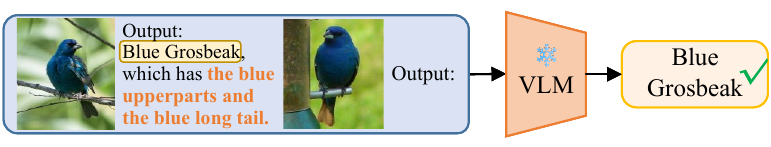}
    \caption{In-context Learning by VDE.}
    \label{fig:intro-c}
  \end{subfigure}
   \begin{subfigure}{\linewidth}
    \centering
    \includegraphics[width=\linewidth]{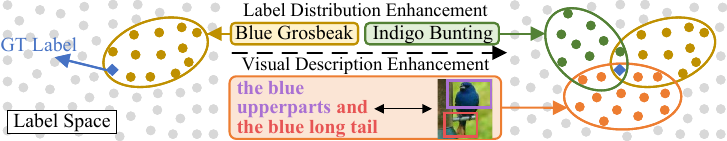}
    \caption{Manipulation of Label Space.}
    \label{fig:intro-d}
  \end{subfigure}
  \caption{(a) Traditional in-context learning with single labels may fail to capture the correct ground truth label. (b) Our Label Distribution Enhancement (LDE) addresses this by considering a broader label distribution. (c) Visual Description Enhancement (VDE) further refines the accurate representation by incorporating detailed visual descriptors. (d) The combined manipulation of the label space with LDE and VDE enables a more precise and comprehensive classification.}
   \label{fig:intro}
\end{figure}    
\section{Introduction}
\label{sec:intro}


Image classification stands as a cornerstone in computer vision and attracts significant research attention due to its critical importance, not only forms the basis for various visual tasks but also serves as a crucial benchmark for diverse network architectures and learning settings, and insights gained from classification studies are instrumental in addressing other vision tasks in these settings.

In the early stages, classification networks predominantly learned from large datasets with human-annotated labels under supervised settings, with ImageNet~\cite{deng2009imagenet} being a notable example. Inspired by the observation that a mature human usually only requires a few or even zero-shot data samples for learning new concepts, zero-shot~\cite{lampert2013attribute} and few-shot~\cite{fei2006one, vinyals2016matching} settings are proposed to make a pre-trained network quickly learn a new class with only a few or even zero-shot data samples. However, a significant limitation of these studies is their reliance on the constrained label space of ImageNet~\cite{Huh_Agrawal_Efros_2016, YuilleLiuGradient2019DeepLearningVision}, hindering the learning of novel labels outside this dataset.

However, the introduction of contrastive learning models like CLIP~\cite{radford2021learning} marked a paradigm shift in zero-shot classification research. After training a two-branch network by contrasting about 400M image-caption pairs, CLIP demonstrates remarkable generalization to concepts not explicitly presented during training. Such zero-shot learning ability, initially observed in GPT-2~\cite{radford2019language} and further developed in GPT-3~\cite{brown2020language}, exemplifies in-context learning (ICL), where the model generalizes to new tasks with minimal examples. \ie, given a few demonstrations to show how tasks should be solved, \eg, the English-Chinese sentence pairs, GPT-3 can capture how to translate English into Chinese without parameter updating. This shift from zero-shot to few-shot learning offers significant potential.

Similarly, in the vision domain, foundation models also evolve from the ones that can achieve prompting engineering to ICL, \eg, Flamingo~\cite{alayrac2022flamingo} can achieve the diverse vision or vision-language tasks through ICL. However, compared with CLIP, VLMs still lag behind in classification tasks since image classification can be viewed as a specific instance of text-image retrieval as the training objective of CLIP~\cite{radford2021learning}. In contrast, VLMs like Flamingo~\cite{alayrac2022flamingo} are trained using a language modeling approach, possibly accounting for the observed performance gap in classification tasks. It is worth noting that existing work on ICL has overlooked the task of bridging the gap in image classification between these two types of models. Addressing this research void, our work delves into enhancing ICL classification performance.

One common approach is to incorporate more samples as in-context examples (ICE). However, this method faces scalability issues. Selecting a greater number of examples from the support set, particularly images, can substantially increase the length of the in-context sequence more than adding text. This increase not only escalates computational demands but also, if the sequence length exceeds that used in training, it could precipitate generalization issues, as highlighted by Press \etal~\cite{press2021train}, leading to a decline in ICL performance. Moreover, while demonstrations in ICL typically aim to refine the label space, the basic single-label format of original ICEs offers limited information. Furthermore, misleading demonstrations could negatively impact performance, similar to the effect of noisy labels in traditional supervised learning.

To address these challenges and enhance performance with fewer shots, we propose two methods for manipulating the label space: Label Distribution Enhancement (LDE) and Visual Description Enhancement (VDE).

LDE, drawing from Label Distribution Learning~\cite{geng2016label,wang2019classification, wang2019theoretical}, posits that an image may associate with multiple labels to varying degrees. We enhance the single-label format of in-context examples (ICE) with richer content, improving ICE efficiency. We consider two strategies: First, manipulating the label embedding was explored, but due to constraints related to dictionary tokenization in VLMs text embedding, \eg, 'sparrow' gets tokenized into 'sp', 'arr', and 'ow'. This approach did not yield optimal results. To overcome this, we directly modify the prompt texts about labels, \eg, calculating the label distribution based on the image similarities and using this distribution in label prompts. Thus providing richer knowledge even when the original in-context label is not a perfect match.

Manipulating the label space in the textual domain typically produces satisfactory results for categories that the model is likely to have encountered. However, this approach faces challenges with rarely encountered labels, especially the labels that do not provide useful visual information like the color or size, \eg, the categories of the cars cannot, thus it is hard to get label distributions by visual similarities. To overcome these limitations, we introduce Visual Description Enhancement (VDE), \ie, for each label, we pair it with the most relevant images and query the model to obtain the detailed visual descriptions for that label. By incorporating these descriptions into the in-context examples, the generalizability and interpretability can be improved. 

Our experiments extend beyond ImageNet to several fine-grained classification datasets. Fortunately, we observed that the intricate relationships between labels in some datasets significantly enhance the effectiveness of our proposed methods. With this work, we aim to provide comprehensive insights into improving ICL performance in image classification tasks and laying the groundwork for future innovations in In-Context Classification. Our contributions are summarized as follows:

\begin{itemize}
    \item We point out the performance disparity between VLMs and contrastive learning paradigms in image classification, being the first work to attempt to bridge this gap by manipulating the label space.
    \item We propose two methods, Label Distribution Enhancement (Section~\ref{subsec:label_dis}) and Visual Description Enhancement (Section~\ref{subsec:description}), for label space manipulation focusing on textual and visual aspects respectively. These methods do not require additional pre-training or fine-tuning, enhancing the generalizability and interpretability of VLMs in in-context classification tasks.
    \item Through extensive experiments, our research showcases notable enhancements in accuracy across various datasets, including gains of up to 6.58\% on ImageNet and considerable improvements ranging from 1.56\% to 20.19\% on three fine-grained datasets under 1-shot conditions with RICES when employing an ensemble of LDE and VDE compared to single label. More details are presented in Section~\ref{subsec:result}.
\end{itemize}

\begin{figure*}
  \centering
    \includegraphics[width=0.9\linewidth]{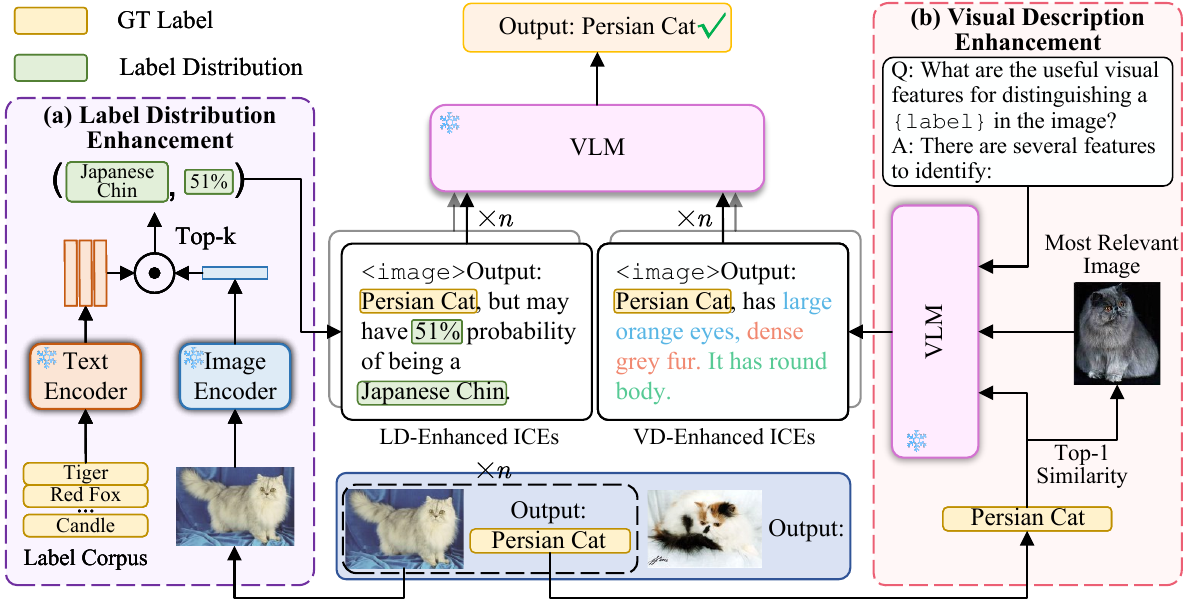}
    \caption{Overview of our proposed Label Space Manipulating, in Label Distribution Enhancement (a), we use image features and text features extracted by CLIP for similarity calculation to get the top-k similar images for a richer label context; In Visual Description Enhancement (b), we query the same VLM with targeted labels and their corresponding most relevant images to generate visual descriptions to assist classification. We explore both textual and visual-focused strategies while the whole VLM parameters are frozen.}
    \label{fig:method}
\end{figure*} 
\section{Related Work}
\label{sec:related}


\noindent\textbf{Image Classification.}
Image classification, a cornerstone of computer vision, has undergone significant evolution. Initially dominated by expansive labeled datasets like ImageNet~\cite{deng2009imagenet} and pioneering models such as AlexNet~\cite{krizhevsky2012imagenet}, the field has progressively shifted towards more versatile approaches. Transfer learning, highlighted in key studies~\cite{mikolov2013distributed,girshick2014rich}, marked a major transition, enabling models to leverage pre-trained features for various tasks. Despite challenges in computational costs and generalization, innovations like zero-shot learning, CLIP~\cite{radford2021learning}, and adaptive CoOp prompting~\cite{zhou2022learning,zhou2022conditional} have paved the way for more adaptive and flexible methodologies. In parallel, few-shot learning, with approaches like prototypical networks~\cite{snell2017prototypical,ye2020heterogeneous}, has emerged, addressing the constraints of data scarcity and further diversifying the image classification toolbox~\cite{Zhang_2020_CVPR,zhang2022deepemd, Zhang_2021_CVPR}. This approach, learning effectively from minimal data, demonstrates the ongoing adaptability and progression in image classification techniques~\cite{Zhang_2021_ICCV, song2022few}.

\noindent\textbf{Prompting Vision-Language Model (VLM).}
While most visual recognition studies hinge on crowd-labeled data for DNN training, often resulting in separate DNNs for each recognition task, this becomes a labor-intensive and protracted paradigm~\cite{zhang2023vision}. Building on the success of NLP with prompt-based techniques, emerging strategies in this domain now employ few-shot learning~\cite{ye2021task,ye2022revisiting}. At the intersection of vision and language, pivotal advancements have streamlined multimodal learning. Unified-IO~\cite{lu2022unified} adeptly handles a broad array of AI challenges by standardizing diverse task inputs into token sequences. Techniques in soft prompting, notably one that reconciles learned and manual prompts, address overfitting concerns~\cite{bulat2022language}. Meanwhile, Prismer leverages domain expertise, requiring minimal training data while achieving competitive results~\cite{liu2023prismer}. Also, pretrained Language Models (PLMs) leverage task-specific prompts to enhance comprehension~\cite{brown2020language,han2021pre, Schick_Schütze_2021}, particularly GPT-3, demonstrate effectiveness in various domains using prompts, underscoring the strengths of ICL. Concurrently, the booming field of multimodal research has begun exploring ICL within VLMs, broadening the scope of applicability of these advanced learning paradigms. VLMs like Flamingo~\cite{alayrac2022flamingo} and Otter~\cite{li2023otter} offer rapid adaptability across tasks with scant annotations, effectively enhancing and highlighting the extensive potential of ICL in the vision-language domain. A recent investigation~\cite{yang2023exploring} has probed into subtle in-context configurations to enhance prompt generation for Image Captioning using VLMs. Despite these advancements, nuanced modifications enhancing in-context visual classification remain underexplored.

\section{Method}
\label{sec:method}

In-context classification within Vision-Language Models (VLMs) can be construed as a conditional text generation task. Given the multi-modal in-context sequence 
$\mathcal{\textbf{\textit{S}}}=\{({\textbf{\textit{I}}}_1,{\textbf{\textit{L}}}_1);({\textbf{\textit{I}}}_2,{\textbf{\textit{L}}}_2);\dots;({\textbf{\textit{I}}}_n,{\textbf{\textit{L}}}_n);\hat{{\textbf{\textit{I}}}}\}$ comprised of $n$-shot image-label pairs $({\textbf{\textit{I}}},{\textbf{\textit{L}}})$ and one test image $\hat{\textbf{\textit{I}}}$, the goal is to fill in the templated sentence, where the slot is filled in by sampling from the probabilistic distribution of potential class $\hat{\textbf{\textit{C}}}=\{\hat{w}_1,\dots,\hat{w}_T\}$ in an auto-regressive manner. Here the $t$-th word $\hat{w}_t$ is sampled from the following word distribution:
\begin{equation} \label{equ:word_distribution}
    P\left(\hat{w}_t\mid \mathcal{\textbf{\textit{S}}},\hat{w}_{1:t-1}\right),
\end{equation}
where the probability $P(\cdot)$ is calculated by a pre-trained Vision-Language Model (VLM) (\eg, Flamingo~\cite{alayrac2022flamingo, awadalla2023openflamingo}). We estimate the score of each test image $\hat{\textbf{\textit{I}}}$ for class $c$ through the log probabilities of the generated words, normalized by the token length, to ensure unbiased class probability outputs across varying label lengths.
\begin{equation} \label{equ:log_probability}
    s(c,\hat{\textbf{\textit{I}}})= \frac{1}{l(c)}\sum_{t=1}^{l(c)} \phi (P(w_t))
\end{equation}
where $l(c)$ is the token length for the class $c$ and $\phi(\cdot)$ is the log probability. (\eg, the class \textit{"Cape\_May\_Warbler"} will be tokenized into \textit{"C, ape, \_, May, \_, War, bler"}, resulting in a token length of 7.)

Prior to manipulating the label space, it is imperative to select appropriate in-context examples (ICEs) from a supporting dataset. We employ two distinct methods for ICE selection: Random Sampling (RS), where $n$ image-label pairs are selected at random from the dataset $\mathcal{D}=\{({\textbf{\textit{I}}}_1,{\textbf{\textit{L}}}_1),\dots,({\textbf{\textit{I}}}_M,{\textbf{\textit{L}}}_M)\}$. To refine the selection quality, we also implement the Retrieval-based In-Context Example Selection (RICES) strategy~\cite{yang2022empirical}, which selects the $n$ most similar images to the test image $\hat{I}$ based on similarity scores.

\subsection{Overall Framework}
\label{subsec:overall}

Although RICES yields substantial improvements over random selection, VLMs still underperform compared to contrastive models like CLIP~\cite{radford2021learning}.

One of the goals of in-context demonstrations is to refine the label space distribution for VLM to better recognize the appearing objects. However, the original single-label In-Context Examples (ICEs) initially provide sparse information, and moreover, when they present misleading demonstrations, this can impair performance akin to the disruptive influence of noisy labels in traditional supervised learning. 

To counteract the limitations of single-label ICEs, we propose two strategies to manipulate the label space as shown in Fig~\ref{fig:method}. Firstly, we enhance the original image-label pairs with label distributions in Section~\ref{subsec:label_dis}, which enriches the classification information and mitigates the negative impact of potentially misleading labels. Secondly, we incorporate visual feature descriptions corresponding to each label in Section~\ref{subsec:description}, enabling a more sophisticated manipulation of the label space at the visual level.

\subsection{Label Distribution Enhancement}
\label{subsec:label_dis}

Addressing the need to enrich the in-context learning experience via label distribution, we initially explore the possibility of adjusting label embeddings. We aim to merge each in-context label embedding with the most similar embeddings from a predefined label set, employing a token-by-token weighted fusion to create a new prompt. This tactic, however, did not produce the anticipated outcomes due to complications with varying token lengths and the intricacies of Byte Pair Encoding (BPE)\cite{shibata1999byte} which is commonly used in VLMs. 

We then shift to direct modification of label prompts, enhancing single labels to label distributions. This involves selecting the most similar labels to the in-context label and calculating their distribution to reflect the relative importance of each label in describing the instance. The cross-modal retrieval capability of CLIP~\cite{radford2021learning} is employed to embed images and labels into a shared space to compute similarities, selecting labels with the highest similarity to the test image. Given $\textbf{\textit{I}}$, we calculate its cross-modal embedding similarities with $\{\textbf{\textit{L}}_1,\dots,{\textbf{\textit{L}}}_M\} \in \mathcal{C}$, and select the labels that have top-$n$ similarities with $\textbf{\textit{I}}$. 

To theoretically analyze the efficacy of LDE, we define the label distribution for each in-context image $\textbf{\textit{I}}$ as:
\begin{equation} \label{equ:lable_space}
    D=\left\{d_{\textbf{\textit{I}}}^{\textbf{\textit{L}}_{1}},d_{\textbf{\textit{I}}}^{\textbf{\textit{L}}_{2}},d_{\textbf{\textit{I}}}^{\textbf{\textit{L}}_{3}},\dots,d_{\textbf{\textit{I}}}^{\textbf{\textit{L}}_{m}}\right\} 
\end{equation}
where $d_{\textbf{\textit{I}}}^{\textbf{\textit{L}}_{m}}$ is derived from the softmax function applied to the similarity scores between the image $\textbf{\textit{I}}$ and $\{\textbf{\textit{L}}_1,\dots,{\textbf{\textit{L}}}_M\} \in \mathcal{C}$, excluding the ground truth label due to our setting of $d_{\textbf{\textit{I}}}^{\textbf{\textit{L}}_{1}}=1$. Thus, we can obtain the label space for Single Label In-Context Examples (ICE) as follows:
\begin{equation} \label{equ:lable_space}
    D=\left\{1,0,0,\dots,0\right\} 
\end{equation}

When the ground truth label in an ICE is insufficient, our proposed label distribution enhancement supplies supplementary data with $d_{\textbf{\textit{I}}}^{\textbf{\textit{L}}_{m}}(m\ge2)$. This addition could potentially offer a suboptimal solution for the model, facilitating more accurate classification. To achieve this objective, we explore three strategies for utilizing label distribution:

(1) Equidistributed Label (EL): Simplifying the distribution to a uniform one by combining the most similar label with the ground truth, implicitly equalizing the importance of each label and providing additional demonstration information.

(2) Distributed Label (DL): Recognizing that the relative importance of different labels varies, we borrowed from label distribution learning (LDL)~\cite{geng2016label} to transform hard labels into a label distribution, reflecting the extent to which each label describes the image. We include similarity scores as pseudo-label probabilities, thus illustrating the relative significance of labels.

(3) Descriptive Distribution (DD): Further, to closely match the token-by-token training and inference approach of VLMs, we attempt to add more descriptive prompts including probability distributions, enhancing both model recognition and interpretability.

\subsection{Visual Descriptions Enhancement}
\label{subsec:description}

Manipulating the label space on the textual front typically yields satisfactory results for categories frequently encountered in conjunction with images on the internet. For instance, the label "bull" is often associated with images of bulls online. However, for labels that the model is unlikely to have encountered, such as labels in Stanford Cars, that do not implicitly suggest visual information, performance bottlenecks occur. Drawing inspiration from the human ability to learn new classes from just a few images, we aim to harness the cross-modal capabilities of VLMs, utilizing visual information to improve the manipulation of the label space.

\begin{figure}[hbt]
  \centering
  \includegraphics[width=0.96\linewidth]{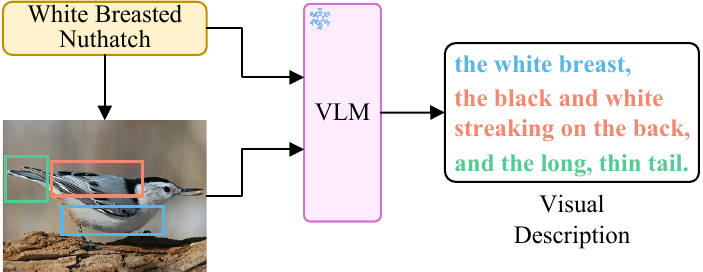}
  \caption{Examples of Visual Description.}
   \label{fig:VDE}
\end{figure}

We enhance in-context labels with detailed visual feature descriptions to provide the model with more granular visual cues for decision-making. To test our method's ability to improve performance without introducing additional model complexity or requiring extra training, we propose an automated process using the same VLMs, such as Open-Flamingo~\cite{awadalla2023openflamingo}. By querying with the prompt "\textit{Q: What are the useful visual features for distinguishing a {label} in the image? A: There are several features to identify:"}, we can elicit unique visual descriptors that effectively correspond to the fine-grained features in the image, as illustrated in Fig~\ref{fig:VDE}.

In a departure from previous efforts~\cite{menon2022visual}, we fully leverage the cross-modal capabilities of VLMs by selecting the most relevant visual features for each class to input into the model. This addition not only provides more effective information but also extrapolates fine-grained features, aiding the VLMs in understanding the characteristics of completely novel objects. Moreover, these visual descriptions allow for clearer differentiation between similar objects, particularly in fine-grained classification datasets, by highlighting distinguished visual features.

We also discover that an ensemble approach combining Label Distribution Enhancement (LDE) and Visual Description Enhancement (VDE) often exceeds the performance of each strategy used separately. This successful manipulation of the label space likely arises because LDE optimizes the textual representation of the label space, providing a probabilistic understanding of label relationships, while VDE contributes detailed visual context, allowing the model to reconcile textual and visual information more cohesively. It enables the model to draw from a more diverse set of cues, thereby enhancing its ability to generalize and accurately classify images.

\section{Experiments}
\label{sec:experiments}

This section of our paper details the implementation aspects of our study. In Section~\ref{subsec:dataset}, we present the datasets utilized, including ImageNet and several fine-grained classification datasets, setting the stage for our experimentation. Section~\ref{subsec:result} explores the outcomes of strategies applied, such as label distribution enhancement in Section~\ref{subsec:lde_perf} and visual feature description in Section~\ref{subsec:des_perf}, alongside a comparison with the default single-label setting in Open Flamingo. Finally, in Section~\ref{subsec:ablation}, we present a comparative analysis with the CLIP as a benchmark in our study due to its efficacy in vision-language tasks.
\begin{table}[h]
\centering
\begin{tabular}{lcccc}
\toprule
 & \textbf{ImageNet} & \textbf{CUB} & \textbf{Cars} & \textbf{Dogs} \\
\midrule
\# of Classes & $1000$ & $200$ & $196$ & $120$ \\
\# of Test Images & $50000$ & $5794$ & $8041$ & $8580$ \\
\bottomrule
\end{tabular}
\caption{Test Dataset Sample Summary}
\label{tab:test_datasets}
\end{table}

\begin{table*}
	\setlength{\tabcolsep}{4pt}
	\centering
	\begin{tabular}{clcccccccccclcccc} 
		\toprule
		\textbf{Dataset}                   &                   & \multicolumn{7}{c}{\textbf{ImageNet}}                            &  & \multicolumn{7}{c}{\textbf{CUB-200}}                              \\ 
		\cline{1-1}\cline{3-9}\cline{11-17}
		\multirow{2}{*}{\textbf{Strategy}} & \multirow{2}{*}{} & \multicolumn{3}{c}{RS}   &  & \multicolumn{3}{c}{RICES} &  & \multicolumn{3}{c}{RS}   &  & \multicolumn{3}{c}{RICES}  \\
		&                   & $ 1 $-shot & $ 2 $-shot & $ 4 $-shot &  & $ 1 $-shot & $ 2 $-shot & $ 4 $-shot  &  & $ 1 $-shot & $ 2 $-shot & $ 4 $-shot &  & $ 1 $-shot & $ 2 $-shot & $ 4 $-shot   \\ 
		\cline{1-1}\cline{3-5}\cline{7-9}\cline{11-13}\cline{15-17}
		SL                        &                   & $ 15.85 $  & $ 21.70 $   & $ 24.31 $  &  & $ 68.50  $  & $ 73.90 $   &$  74.70 $    &  & $ 7.47 $   & $ 10.29 $  & $ 12.44 $  &  & $ 48.86  $ & $ 63.07 $  & $ 59.98 $    \\
		LDE(EL)                   &                   & $ 13.38  $ &$  23.29 $  & $ 25.08 $  &  & $ 70.23 $  &$  75.08  $ & $ \textbf{75.02}  $  &  & $ \textbf{9.30} $    & $ \textbf{12.41}  $ & $ \textbf{13.43}  $ &  &$  58.27  $ & $ \textbf{67.33}  $ & $ \textbf{63.36} $    \\
		LDE(DL)                   &                   & $ 15.75 $  & $ 23.07 $  & $ 25.29 $  &  & $ 73.10 $   & $ 75.25  $ & $ 74.12 $   &  &$  7.53 $   & $ 12.10  $  & $ 12.82 $  &  & $ 64.76 $  & $ 66.40  $  & $ 61.36 $    \\
		LDE(DD)                   &                   & $ \textbf{17.85} $  & $ \textbf{23.85} $  & $ \textbf{25.97} $  &  & $ \textbf{73.63} $  & $ \textbf{75.83}  $ & $ 74.03  $  &  &$  6.39  $  & 11.30   & 12.25  &  & $ \textbf{66.62}  $ & $ 67.22 $  & $ 62.75 $    \\ 
		\hline\hline
		\textbf{Dataset}                   &                   & \multicolumn{7}{c}{\textbf{Stanford Dogs}}                       &  & \multicolumn{7}{c}{\textbf{Standford Cars}}                       \\ 
		\cline{1-1}\cline{3-9}\cline{11-17}
		\multirow{2}{*}{\textbf{Strategy}} & \multirow{2}{*}{} & \multicolumn{3}{c}{RS}   &  & \multicolumn{3}{c}{RICES} &  & \multicolumn{3}{c}{RS}   &  & \multicolumn{3}{c}{RICES}  \\
		&                   & $1$-shot & $2$-shot & $4$-shot &  & $1$-shot & $2$-shot & $4$-shot  &  & $1$-shot & $2$-shot & $4$-shot &  & $1$-shot & $2$-shot & $4$-shot   \\ 
		\cline{1-1}\cline{3-5}\cline{7-9}\cline{11-13}\cline{15-17}
		SL                        &                   & $10.30$   & $21.34$  & $28.21$  &  & $61.40 $  &$ 66.38$  & $\textbf{66.83}$   &  & $41.77  $ & $\textbf{49.56} $  & $\textbf{49.02} $  &  & $74.19 $  & $\textbf{79.55 }$  & $\textbf{74.83 }$    \\
		LDE(EL)                   &                   & $11.40$   & $\textbf{26.60}$   & $\textbf{29.43}$  &  &$ 64.23$  & $\textbf{67.16}$  & $66.06 $  &  & $\textbf{46.33 } $ & $49.46 $  & $48.08 $  &  & $73.14 $  & $77.69 $  & $74.64 $     \\
		LDE(DL)                   &                   & $\textbf{11.15}$  & $26.36$  & $29.22$  &  & $65.54$  & $66.83$  & $64.15$   &  & $45.06 $  & $48.35 $  & $47.26 $  &  & $73.64 $  & $74.97 $  & $72.57 $    \\
		LDE(DD)                   &                   &$ 7.21$   & $25.13 $ & $28.28$  &  & $\textbf{65.55}$  & $66.69 $ & $63.48$   &  & $41.89 $  & $49.38 $  & $48.03 $ &  & $\textbf{75.64} $  & $77.73 $  & $74.02  $   \\
		\bottomrule
	\end{tabular}
	\caption{LDE Results}
	\label{tab:results}
\end{table*}

\subsection{Datasets and Implementation Details}
\label{subsec:dataset}


\noindent\textbf{Datasets.}
To evaluate the proposed strategies, we conduct extensive experimentation on four image classification benchmarks, including both generic and fine-grained object classifications. For generic objects, we adopt ImageNet~\cite{deng2009imagenet} in line with existing research precedents~\cite{awadalla2023openflamingo,alayrac2022flamingo}. For fine-grained object classification, we choose CUB-200~\cite{WelinderEtal2010}, Stanford Dogs~\cite{khosla2011novel}, and Stanford Cars~\cite{krause20133d}, as detailed in the Table~\ref{tab:test_datasets}. These datasets demand a high level of visual detail for accurate categorization. Their fine-grained nature poses considerable challenges in differentiating closely related categories, making them well-suited for assessing our in-context classification approach with Vision-Language Models (VLMs).

It is worth noting that for all the selected datasets, we apply a train/test split. Concretely, the training subset is employed as a support set for selecting ICEs, and the testing subset is reserved for evaluation.


\noindent\textbf{Implementation Details.} 
In our implementation, we utilize the Open-Flamingo-3B-vitl-mpt1b model with ViT-L/14, setting the length penalty at 1.0 and the maximum generation length at 20. We concentrate on 1, 2, and 4 shots, as performance declined beyond 8 shots. Accuracy was the primary metric due to its clarity and common usage. Notably, our experimental setup required no additional training or fine-tuning. All experiments were conducted on a single NVIDIA RTX 3090 GPU with BF16 acceleration.

 

\subsection{Results and Discussion}
\label{subsec:result}

In our experimental efforts to manipulate the label space for In-Context Classification, we implement strategies such as label distribution enhancement and visual feature description. We perform both quantitative and qualitative analyses, comparing these approaches against the baseline single-label setting in Open Flamingo across a variety of classification datasets.

\begin{table*}
\setlength{\tabcolsep}{4pt}
\centering
\begin{tabular}{ccccccccccccccccc} 
\toprule
\textbf{Dataset}                   &                   & \multicolumn{7}{c}{\textbf{ImageNet}}                   &  & \multicolumn{7}{c}{\textbf{CUB-200}}                     \\ 
\cline{1-1}\cline{3-9}\cline{11-17}
\multirow{2}{*}{\textbf{Strategy}} & \multirow{2}{*}{} & \multicolumn{3}{c}{RS}   &  & \multicolumn{3}{c}{RICES} &  & \multicolumn{3}{c}{RS}   &  & \multicolumn{3}{c}{RICES}  \\
                                   &                   & $1$-shot & $2$-shot & $4$-shot &  & $1$-shot & $2$-shot & $4$-shot  &  & $1$-shot & $2$-shot & $4$-shot &  & $1$-shot & $2$-shot & $4$-shot   \\ 
\cline{1-1}\cline{3-5}\cline{7-9}\cline{11-13}\cline{15-17}
SL                                 &                   & $15.85$  & $21.70$  & $24.31$  &  & $68.50$  & $73.90$  & $\textbf{74.70}$   &  & $\textbf{7.47} $  & $10.29$  & $12.44$  &  & $48.86$  & $63.07$  & $59.98$    \\
LDE(DD)                            &                   & $\textbf{17.85}$  & $23.85$  & $25.97$  &  & $73.63$  & $75.83$  & $74.03$   &  & $6.39$   & $11.30$  & $12.25$  &  & $66.62 $ & $67.22$  & $\textbf{62.75}$    \\
VDE                                &                   & $8.47$   & $16.90$  & $21.17 $ &  & $73.65$  & $75.00$  & $70.89$   &  & $2.95$   & $8.01$   & $11.84$  &  &$ 67.97$  & $\textbf{69.05}$  & $58.06$    \\
ENS.                                &                   & $15.38$  & $\textbf{24.04}$  & $\textbf{26.71}$  &  & $\textbf{75.08}$  & $\textbf{76.21}$  & $72.73$   &  & $7.23$   & $\textbf{11.60}$  & $\textbf{13.14}$  &  & $\textbf{69.05}$  & $68.95$  & $60.37$    \\ 
\hline\hline
\textbf{Dataset}                   &                   & \multicolumn{7}{c}{\textbf{Stanford Dogs}}              &  & \multicolumn{7}{c}{\textbf{Standford Cars}}              \\ 
\cline{1-1}\cline{3-9}\cline{11-17}
\multirow{2}{*}{\textbf{Strategy}} & \multirow{2}{*}{} & \multicolumn{3}{c}{RS}   &  & \multicolumn{3}{c}{RICES} &  & \multicolumn{3}{c}{RS}   &  & \multicolumn{3}{c}{RICES}  \\
                                   &                   & 1-shot & 2-shot & 4-shot &  & 1-shot & 2-shot & 4-shot  &  & 1-shot & 2-shot & 4-shot &  & 1-shot & 2-shot & 4-shot   \\ 
\cline{1-1}\cline{3-5}\cline{7-9}\cline{11-13}\cline{15-17}
SL                                 &                   & $\textbf{10.30}$  & $21.34 $ & $28.21$  &  & $61.40$  & $66.38 $ & $\textbf{66.83}$   &  & $41.77$  & $\textbf{49.56}$  & $\textbf{49.02}$  &  & $74.19$  & $79.55$  & $\textbf{74.83}$    \\
LDE(DD)                            &                   & $7.21$   & $\textbf{25.13}$  & $28.28$  &  & $\textbf{65.55}$  & $66.69$  & $63.48$   &  & $\textbf{41.89 }$  & $49.38 $  & $48.03 $ &  & $75.64 $  & $77.73 $  & $74.02$    \\
VDE                                &                   & $4.08$   & $11.61$  & $21.14 $ &  & $63.87$  & $64.31$  & $55.58$   &  & $32.04 $   & $46.70 $  & $47.31 $  &  & $\textbf{79.14 }$  & $\textbf{80.57 }$  & $72.04 $    \\
ENS.                                &                   & $8.22$  & $24.86$  & $\textbf{29.52}$  &  & $62.96 $ & $\textbf{67.76}$  & $61.54$   &  & $37.25$  & $49.52$  & $48.91$  &  & $78.88$  & $80.11$  & $73.32$    \\
\bottomrule
\end{tabular}
\caption{VDE Results, \textbf{"ENS."} denotes an experimental ensemble of \textbf{VDE} with \textbf{LDE (DD)}, representing a combined approach that integrates both visual and linguistic description enhancements.}
\label{tab:results2}
\end{table*}
\subsubsection{Performance of LDE}
\label{subsec:lde_perf}

Table~\ref{tab:results} outlines the comparative performances of various methodologies evaluated in this study.

\noindent\textbf{In-Context Examples Matter.} In our experimental framework, we explore two distinct methodologies (RICES and RS) for selecting In-Context Examples (ICE). We find that both the quality and quantity of ICE significantly impact the results. This becomes especially apparent when we analyze performances across various datasets. Using the standard \textbf{single-label (SL)} approach, we observe a marked improvement when moving from 1-shot to 2-shot, underscoring the influence of the number of ICE. Moreover, comparing RS and RICES, we note that RICES—by choosing images more akin to the test examples—offers higher quality ICE than the random selection method employed by RS. These observations underscore the pivotal role of ICE selection in classification performance. They demonstrate that strategically selecting high-quality ICE, in conjunction with determining their optimal number, is crucial for superior classification outcomes.

\noindent\textbf{Effects of LDE Strategies.} Our analysis, as illustrated in~\ref{tab:results}, commences with a comparison between LDE strategies and the standard SL approach. Across four datasets except for Stanford Cars, LDE strategies, particularly LDE (DD) and LDE (EL) under the RICES retrieval method, consistently outperform the SL approach. A notable observation is the significant superiority of LDE strategies in 1-shot scenarios, where they distinctly surpass SL in terms of performance. This underscores the effectiveness of LDE in contexts with minimal data. Delving into specific results, on Stanford Cars, ImageNet, CUB-200, and Stanford Dogs, the performance margins achieved by LDE strategies are \textbf{75.64\%} in the 1-shot setting, \textbf{75.83\%}, \textbf{67.33\%} in the 2-shot setting, and \textbf{66.83\%} in the 4-shot setting, respectively. For LDE, its integration of richer semantic information remarkably enhances the recognition capabilities and interpretability of the VLM model. However, in the context of Stanford Cars, it is observed that most LDE strategies do not surpass the SL approach. Based on our observations and experiments, presumably, classification using generative models and LDE strategies tends to confound the models for classes that differ very little from each other. (e.g., \textit{\textbf{"Acura Integra Type R 2001"}} and \textit{\textbf{"Acura Integra Type S 2024"}}) In another strategy, VDE, providing visual information significantly improves this classification~\ref{subsec:des_perf}  Moreover, an interesting pattern emerges when examining the performance variation with the number of shots. In few-shot scenarios, such as 1 or 2 shots, LDE (DD) demonstrates superior efficacy. Yet, as the shot number increases, we observe a plateau in its performance. This trend suggests a potential constraint related to prompt length and model capacity. To address this bottleneck, employing models with larger parameter sets could be a potential solution. These insights reinforce the effectiveness of LDE in improving classification accuracy in complex label spaces and highlight areas for future enhancements.

\noindent\textbf{More efficient utilization of ICE of LDE Over SL.} When applying LDE strategies under the RICES retrieval method, our analysis reveals a clear trend regarding the effectiveness of different shot numbers. Across various datasets, we find that two shots often yield better results than four shots in the SL approach. This pattern is particularly pronounced in CUB-200, where even a single shot using LDE(DL) and LDE(DD) outperforms two shots, indicating that our method's efficiency allows one high-quality image to match the impact of two. This discovery underscores the importance of label distribution in boosting the performance of VLMs for complex classification tasks.

\subsubsection{Performance of VDE}
\label{subsec:des_perf}
Table~\ref{tab:results2} is showcased to illustrate the differential impacts of Visual Description Enhancement (VDE) techniques in our study. 

\noindent\textbf{Performance of VDE across dataset types.} It is important to note the effectiveness of Visual Description Enhancement (VDE) varies across different types of datasets.  While VDE significantly improves over the SL approach in fine-grained datasets, its performance is not consistently superior to LDE in more generalized datasets like ImageNet, with a marginal improvement of 0.02\% in the 1-shot scenario. Additionally, the quality and insertion order of descriptions play a crucial role in outcomes, as detailed in the appendix. Despite this, the overall efficacy of VDE, particularly in fine-grained datasets like Stanford Cars and CUB-200, where VDE outperformed the SL approach by 14.08\% in 2-shot and 19.11\% in 1-shot scenarios, underscores its transformative potential in enhancing ICL. In the case of the Stanford Dogs, our preliminary assessment suggests that due to the substantial variation in coat texture and color among dogs of the same species, the impact of the VDE approach might be less pronounced. Our rigorous testing and evaluation further highlight this potential, suggesting that the utility of VDE is maximized in tasks demanding intricate classification capabilities.

\noindent\textbf{Advantages of VDE in In-Context Classification.} Our findings suggest that augmenting in-context examples with rich visual descriptions imparts a discernible advantage. This is especially noticeable in datasets with subtle class distinctions, like Stanford Cars, where such enhancements enable LLMs to draw upon a more vivid tableau of contextual cues (\eg, \textit{\textbf{BMW X5 2007: "a black body, a silver bumper, a black grille, a large chrome headlight"}}), further detailed descriptions are provided in the supplementary materials. This leads to more refined and discerning classification capabilities.

\noindent\textbf{Ensemble of VDE and LDE. (ENS.)} In our experiments, we try to combine LDE (DD) and VDE for better performance, we obtain test images and compute the average of the logarithms of class outputs after applying the two enhancement strategies mentioned above to get the final classification results. This combination is more effective on certain datasets. Scores greater than their respective stacked parts are achieved on all four datasets, with the best on ImageNet being 1.43\% higher, CUB-200 1.08\%, Standford Dogs 1.66\%, as shown in Table~\ref{tab:results2}.

\begin{figure}[t]
  \centering
  \includegraphics[width=\linewidth]{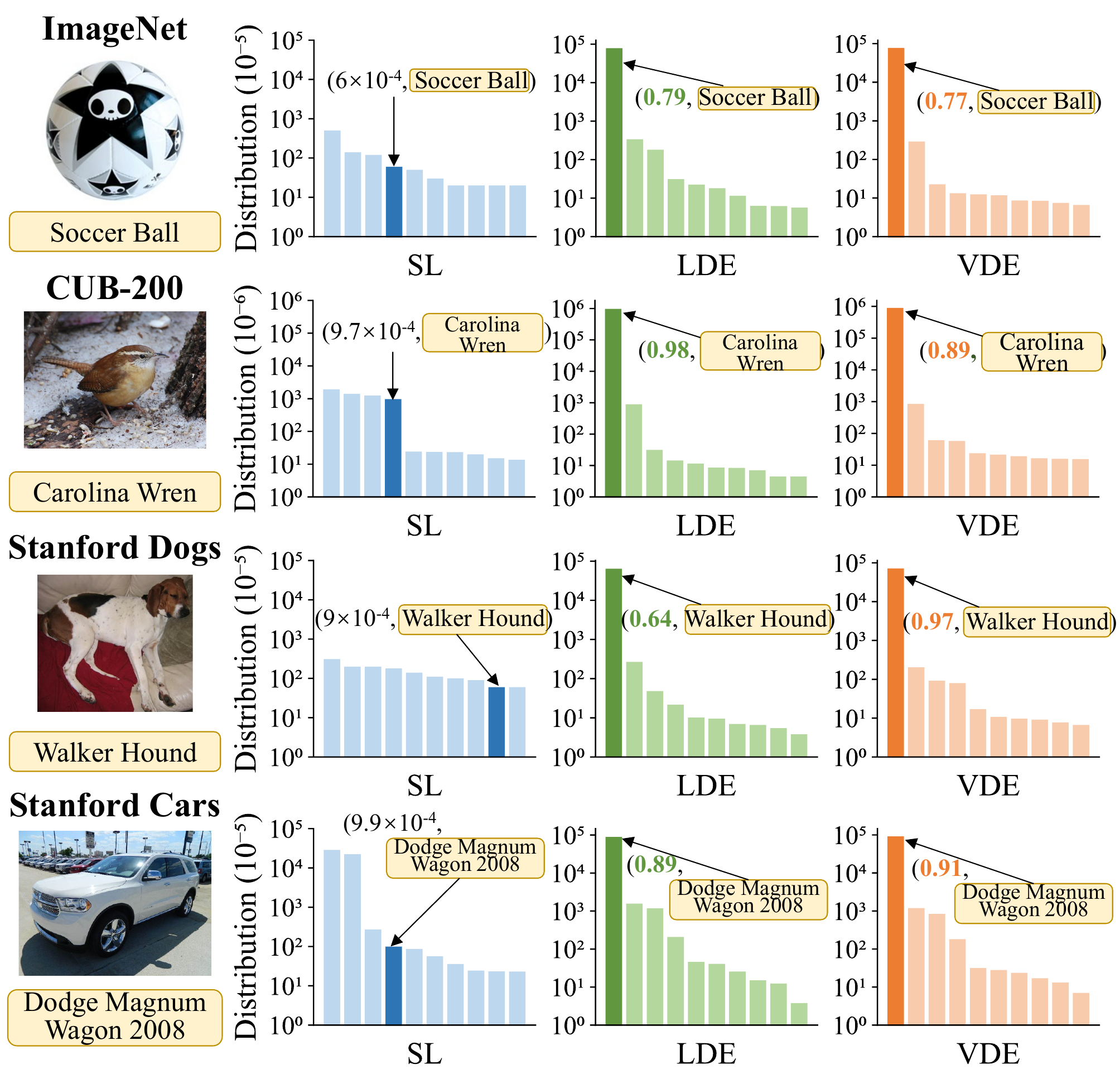}
  \caption{This is a visualization of our proposed method that eliminates some of the confusing situations in normal classification. Our method correctly classifies the category that was originally misclassified as soccer helmets as footballs, and it is intuitively clear that the probability of this category is significantly higher than that of the other categories both in label distribution and visual description.}
   \label{fig:visualization}
\end{figure}

\subsubsection{Deeper Visualization}
\label{subsec:visual}
Fig.~\ref{fig:visualization} illustrates how label manipulations affect the output distribution of the VLM. Under the SL strategy, the prediction of the VLM often appears perplexing, possibly due to the lack of richness and diversity in label information. By contrast, our approach enriches the label space with more detailed and varied information, which not only enhances the understanding of the VLM of each category but also boosts its confidence in making precise label predictions. Consequently, this enriched label information indirectly alters the label space, playing a pivotal role in improving the performance of in-context classification, particularly in scenarios demanding a nuanced understanding of complex categories.

\subsection{Comparison with CLIP}
\label{subsec:ablation}
We compare our strategies with the CLIP\footnote{The evaluation results for CLIP are replicated using the open-clip repository, achieving identical outcomes on ImageNet as originally reported.} model, which is employed as the vision encoder in the Open Flamingo setup, under identical conditions, offering a crucial benchmark against a recognized standard and providing context for evaluating our performance.

While our approach surpasses CLIP in certain datasets, it does not in Standford Dogs, highlighting opportunities for exploration. Specifically, the robust feature extraction of CLIP provides strong baseline results, but it falls short in fine-grained classifications where our method excels. Utilizing a 3B Open-Flamingo model for VLM-based classification, our method outperforms CLIP in datasets like ImageNet, CUB-200, and Stanford Cars, with improvements of 0.67\%, 12.15\%, and 3.38\%, respectively in a 2-shot scenario, as shown in Table~\ref{tab:results3}. Specifically, in CUB-200, where classification is often constrained by specific bird names, sizes, and colors, CLIP struggles to effectively align images with textual descriptions because its performance only has 56.90\% accuracy. This limitation, as exemplified in the above section, means that our approach effectively bridges this gap, leveraging the detailed nuances that CLIP misses, thereby enhancing classification accuracy in such contextually rich scenarios. These results, achieved without additional training or fine-tuning, underscore the efficiency of our fine-grained classification strategy, which leverages visual descriptors and label manipulation for high precision and detailed differentiation, proving effective in contexts requiring nuanced understanding.

\definecolor{Gray}{rgb}{0.501,0.501,0.501}
\definecolor{GuardsmanRed}{rgb}{0.815,0,0}
\definecolor{MilanoRed}{rgb}{0.717,0.05,0.05}
\begin{table}
\centering
\small
\setlength{\tabcolsep}{2pt}
\begin{tblr}{
  cells = {c},
  column{2} = {fg=Gray},
  column{3} = {fg=black},
  column{5} = {fg=GuardsmanRed},
  cell{2}{5} = {fg=MilanoRed},
  cell{4}{5} = {fg=Gray},
  hline{1,6} = {-}{0.08em},
  hline{2} = {-}{},
}
\textbf{Dataset} & \textbf{SL} & \textbf{CLIP} & \textbf{ENS.} & $\Delta$ \\
ImageNet         & $74.70$     & $75.54$       & $76.21$               & $0.67$   \\
CUB-200          & $63.07$     & $56.90$       & $69.05$               & $12.15$  \\
Stanford Dogs    & $66.83$     & $69.09$       & $67.76$               & $-1.33$  \\
Stanford Cars    & $79.55$     & $77.19$       & $80.11$               & $2.92$  
\end{tblr}
\caption{Comparing our certain methods under 2-shot scenario with CLIP, $\Delta$ means gap between ENS. 2-shot and CLIP.}
\label{tab:results3}
\end{table}
\section{Conclusion and Limitations}
\label{sec:conclusion}

In this study, we introduce a novel approach for in-context classification in Vision-Language Models (VLMs), employing two complementary methods: Label Distribution Enhancement (LDE) and Visual Description Enhancement (VDE). Our experiments show that a combination of LDE and VDE often outperforms each method individually, highlighting their potential in enhancing VLMs in-context learning capabilities without the need for additional training.

However, our research has notable limitations. We conduct experiments on a smaller-scale VLM with a limited number of shots, limiting our insights into the scalability and effectiveness of our methods in larger or more complex models. Additionally, our strategies focus solely on prompt manipulation, excluding updates to model parameters or embedding adjustments. We plan to explore these aspects in the future.
\newpage
{
    \small
    \bibliographystyle{ieeenat_fullname}
    \bibliography{bibliography}

\begin{thebibliography}{43}
\providecommand{\natexlab}[1]{#1}
\providecommand{\url}[1]{\texttt{#1}}
\expandafter\ifx\csname urlstyle\endcsname\relax
  \providecommand{\doi}[1]{doi: #1}\else
  \providecommand{\doi}{doi: \begingroup \urlstyle{rm}\Url}\fi

\bibitem[Alayrac et~al.(2022)Alayrac, Donahue, Luc, Miech, Barr, Hasson, Lenc, Mensch, Millican, Reynolds, Ring, Rutherford, Cabi, Han, Gong, Samangooei, Monteiro, Menick, Borgeaud, Brock, Nematzadeh, Sharifzadeh, Binkowski, Barreira, Vinyals, Zisserman, and Simonyan]{alayrac2022flamingo}
Jean-Baptiste Alayrac, Jeff Donahue, Pauline Luc, Antoine Miech, Iain Barr, Yana Hasson, Karel Lenc, Arthur Mensch, Katie Millican, Malcolm Reynolds, Roman Ring, Eliza Rutherford, Serkan Cabi, Tengda Han, Zhitao Gong, Sina Samangooei, Marianne Monteiro, Jacob Menick, Sebastian Borgeaud, Andrew Brock, Aida Nematzadeh, Sahand Sharifzadeh, Mikolaj Binkowski, Ricardo Barreira, Oriol Vinyals, Andrew Zisserman, and Karen Simonyan.
\newblock Flamingo: a visual language model for few-shot learning, 2022.

\bibitem[Awadalla et~al.(2023)Awadalla, Gao, Gardner, Hessel, Hanafy, Zhu, Marathe, Bitton, Gadre, Sagawa, Jitsev, Kornblith, Koh, Ilharco, Wortsman, and Schmidt]{awadalla2023openflamingo}
Anas Awadalla, Irena Gao, Josh Gardner, Jack Hessel, Yusuf Hanafy, Wanrong Zhu, Kalyani Marathe, Yonatan Bitton, Samir Gadre, Shiori Sagawa, Jenia Jitsev, Simon Kornblith, Pang~Wei Koh, Gabriel Ilharco, Mitchell Wortsman, and Ludwig Schmidt.
\newblock Openflamingo: An open-source framework for training large autoregressive vision-language models, 2023.

\bibitem[Brown et~al.(2020)Brown, Mann, Ryder, Subbiah, Kaplan, Dhariwal, Neelakantan, Shyam, Sastry, Askell, et~al.]{brown2020language}
Tom Brown, Benjamin Mann, Nick Ryder, Melanie Subbiah, Jared~D Kaplan, Prafulla Dhariwal, Arvind Neelakantan, Pranav Shyam, Girish Sastry, Amanda Askell, et~al.
\newblock Language models are few-shot learners.
\newblock \emph{Advances in neural information processing systems}, 33:\penalty0 1877--1901, 2020.

\bibitem[Bulat and Tzimiropoulos(2022)]{bulat2022language}
Adrian Bulat and Georgios Tzimiropoulos.
\newblock Language-aware soft prompting for vision \& language foundation models.
\newblock \emph{arXiv preprint arXiv:2210.01115}, 2022.

\bibitem[Deng et~al.(2009)Deng, Dong, Socher, Li, Li, and Fei-Fei]{deng2009imagenet}
Jia Deng, Wei Dong, Richard Socher, Li-Jia Li, Kai Li, and Li Fei-Fei.
\newblock Imagenet: A large-scale hierarchical image database.
\newblock In \emph{2009 IEEE conference on computer vision and pattern recognition}, pages 248--255. Ieee, 2009.

\bibitem[Fei-Fei et~al.(2006)Fei-Fei, Fergus, and Perona]{fei2006one}
Li Fei-Fei, Robert Fergus, and Pietro Perona.
\newblock One-shot learning of object categories.
\newblock \emph{IEEE transactions on pattern analysis and machine intelligence}, 28\penalty0 (4):\penalty0 594--611, 2006.

\bibitem[Geng(2016)]{geng2016label}
Xin Geng.
\newblock Label distribution learning.
\newblock \emph{IEEE Transactions on Knowledge and Data Engineering}, 28\penalty0 (7):\penalty0 1734--1748, 2016.

\bibitem[Girshick et~al.(2014)Girshick, Donahue, Darrell, and Malik]{girshick2014rich}
Ross Girshick, Jeff Donahue, Trevor Darrell, and Jitendra Malik.
\newblock Rich feature hierarchies for accurate object detection and semantic segmentation.
\newblock In \emph{Proceedings of the IEEE conference on computer vision and pattern recognition}, pages 580--587, 2014.

\bibitem[Han et~al.(2021)Han, Zhang, Ding, Gu, Liu, Huo, Qiu, Yao, Zhang, Zhang, et~al.]{han2021pre}
Xu Han, Zhengyan Zhang, Ning Ding, Yuxian Gu, Xiao Liu, Yuqi Huo, Jiezhong Qiu, Yuan Yao, Ao Zhang, Liang Zhang, et~al.
\newblock Pre-trained models: Past, present and future.
\newblock \emph{AI Open}, 2:\penalty0 225--250, 2021.

\bibitem[Huh et~al.(2016)Huh, Agrawal, and Efros]{Huh_Agrawal_Efros_2016}
Minyoung Huh, Pulkit Agrawal, and AlexeiA. Efros.
\newblock What makes imagenet good for transfer learning.
\newblock \emph{arXiv: Computer Vision and Pattern Recognition,arXiv: Computer Vision and Pattern Recognition}, 2016.

\bibitem[Khosla et~al.(2011)Khosla, Jayadevaprakash, Yao, and Li]{khosla2011novel}
Aditya Khosla, Nityananda Jayadevaprakash, Bangpeng Yao, and Fei-Fei Li.
\newblock Novel dataset for fine-grained image categorization: Stanford dogs.
\newblock In \emph{Proc. CVPR workshop on fine-grained visual categorization (FGVC)}. Citeseer, 2011.

\bibitem[Krause et~al.(2013)Krause, Stark, Deng, and Fei-Fei]{krause20133d}
Jonathan Krause, Michael Stark, Jia Deng, and Li Fei-Fei.
\newblock 3d object representations for fine-grained categorization.
\newblock In \emph{Proceedings of the IEEE international conference on computer vision workshops}, pages 554--561, 2013.

\bibitem[Krizhevsky et~al.(2012)Krizhevsky, Sutskever, and Hinton]{krizhevsky2012imagenet}
Alex Krizhevsky, Ilya Sutskever, and Geoffrey~E Hinton.
\newblock Imagenet classification with deep convolutional neural networks.
\newblock \emph{Advances in neural information processing systems}, 25, 2012.

\bibitem[Lampert et~al.(2013)Lampert, Nickisch, and Harmeling]{lampert2013attribute}
Christoph~H Lampert, Hannes Nickisch, and Stefan Harmeling.
\newblock Attribute-based classification for zero-shot visual object categorization.
\newblock \emph{IEEE transactions on pattern analysis and machine intelligence}, 36\penalty0 (3):\penalty0 453--465, 2013.

\bibitem[Li et~al.(2023)Li, Zhang, Chen, Wang, Yang, and Liu]{li2023otter}
Bo Li, Yuanhan Zhang, Liangyu Chen, Jinghao Wang, Jingkang Yang, and Ziwei Liu.
\newblock Otter: A multi-modal model with in-context instruction tuning.
\newblock \emph{arXiv preprint arXiv:2305.03726}, 2023.

\bibitem[Liu et~al.(2023)Liu, Fan, Johns, Yu, Xiao, and Anandkumar]{liu2023prismer}
Shikun Liu, Linxi Fan, Edward Johns, Zhiding Yu, Chaowei Xiao, and Anima Anandkumar.
\newblock Prismer: A vision-language model with an ensemble of experts.
\newblock \emph{arXiv preprint arXiv:2303.02506}, 2023.

\bibitem[Lu et~al.(2022)Lu, Clark, Zellers, Mottaghi, and Kembhavi]{lu2022unified}
Jiasen Lu, Christopher Clark, Rowan Zellers, Roozbeh Mottaghi, and Aniruddha Kembhavi.
\newblock Unified-io: A unified model for vision, language, and multi-modal tasks.
\newblock \emph{arXiv preprint arXiv:2206.08916}, 2022.

\bibitem[Menon and Vondrick(2022)]{menon2022visual}
Sachit Menon and Carl Vondrick.
\newblock Visual classification via description from large language models.
\newblock \emph{arXiv preprint arXiv:2210.07183}, 2022.

\bibitem[Mikolov et~al.(2013)Mikolov, Sutskever, Chen, Corrado, and Dean]{mikolov2013distributed}
Tomas Mikolov, Ilya Sutskever, Kai Chen, Greg~S Corrado, and Jeff Dean.
\newblock Distributed representations of words and phrases and their compositionality.
\newblock \emph{Advances in neural information processing systems}, 26, 2013.

\bibitem[Press et~al.(2021)Press, Smith, and Lewis]{press2021train}
Ofir Press, Noah~A Smith, and Mike Lewis.
\newblock Train short, test long: Attention with linear biases enables input length extrapolation.
\newblock \emph{arXiv preprint arXiv:2108.12409}, 2021.

\bibitem[Radford et~al.(2019)Radford, Wu, Child, Luan, Amodei, Sutskever, et~al.]{radford2019language}
Alec Radford, Jeffrey Wu, Rewon Child, David Luan, Dario Amodei, Ilya Sutskever, et~al.
\newblock Language models are unsupervised multitask learners.
\newblock \emph{OpenAI blog}, 1\penalty0 (8):\penalty0 9, 2019.

\bibitem[Radford et~al.(2021)Radford, Kim, Hallacy, Ramesh, Goh, Agarwal, Sastry, Askell, Mishkin, Clark, et~al.]{radford2021learning}
Alec Radford, Jong~Wook Kim, Chris Hallacy, Aditya Ramesh, Gabriel Goh, Sandhini Agarwal, Girish Sastry, Amanda Askell, Pamela Mishkin, Jack Clark, et~al.
\newblock Learning transferable visual models from natural language supervision.
\newblock In \emph{International conference on machine learning}, pages 8748--8763. PMLR, 2021.

\bibitem[Schick and Schütze(2021)]{Schick_Schütze_2021}
Timo Schick and Hinrich Schütze.
\newblock Exploiting cloze questions for few shot text classification and natural language inference.
\newblock In \emph{Proceedings of the 16th Conference of the European Chapter of the Association for Computational Linguistics: Main Volume}, 2021.

\bibitem[Shibata et~al.(1999)Shibata, Kida, Fukamachi, Takeda, Shinohara, Shinohara, and Arikawa]{shibata1999byte}
Yusuxke Shibata, Takuya Kida, Shuichi Fukamachi, Masayuki Takeda, Ayumi Shinohara, Takeshi Shinohara, and Setsuo Arikawa.
\newblock Byte pair encoding: A text compression scheme that accelerates pattern matching.
\newblock 1999.

\bibitem[Snell et~al.(2017)Snell, Swersky, and Zemel]{snell2017prototypical}
Jake Snell, Kevin Swersky, and Richard Zemel.
\newblock Prototypical networks for few-shot learning.
\newblock \emph{Advances in neural information processing systems}, 30, 2017.

\bibitem[Song et~al.(2022)Song, Zhang, and Lin]{song2022few}
Nan Song, Chi Zhang, and Guosheng Lin.
\newblock Few-shot open-set recognition using background as unknowns.
\newblock In \emph{Proceedings of the 30th ACM International Conference on Multimedia}, pages 5970--5979, 2022.

\bibitem[Vinyals et~al.(2016)Vinyals, Blundell, Lillicrap, Wierstra, et~al.]{vinyals2016matching}
Oriol Vinyals, Charles Blundell, Timothy Lillicrap, Daan Wierstra, et~al.
\newblock Matching networks for one shot learning.
\newblock \emph{Advances in neural information processing systems}, 29, 2016.

\bibitem[Wang and Geng(2019{\natexlab{a}})]{wang2019classification}
Jing Wang and Xin Geng.
\newblock Classification with label distribution learning.
\newblock In \emph{Proceedings of the Twenty-Eighth International Joint Conference on Artificial Intelligence}, 2019{\natexlab{a}}.

\bibitem[Wang and Geng(2019{\natexlab{b}})]{wang2019theoretical}
Jing Wang and Xin Geng.
\newblock Theoretical analysis of label distribution learning.
\newblock In \emph{Proceedings of the AAAI Conference on Artificial Intelligence}, pages 5256--5263, 2019{\natexlab{b}}.

\bibitem[Welinder et~al.(2010)Welinder, Branson, Mita, Wah, Schroff, Belongie, and Perona]{WelinderEtal2010}
P. Welinder, S. Branson, T. Mita, C. Wah, F. Schroff, S. Belongie, and P. Perona.
\newblock {Caltech-UCSD Birds 200}.
\newblock Technical Report CNS-TR-2010-001, California Institute of Technology, 2010.

\bibitem[Yang et~al.(2023)Yang, Wu, Yang, Chen, and Geng]{yang2023exploring}
Xu Yang, Yongliang Wu, Mingzhuo Yang, Haokun Chen, and Xin Geng.
\newblock Exploring diverse in-context configurations for image captioning.
\newblock In \emph{Thirty-seventh Conference on Neural Information Processing Systems}, 2023.

\bibitem[Yang et~al.(2022)Yang, Gan, Wang, Hu, Lu, Liu, and Wang]{yang2022empirical}
Zhengyuan Yang, Zhe Gan, Jianfeng Wang, Xiaowei Hu, Yumao Lu, Zicheng Liu, and Lijuan Wang.
\newblock An empirical study of gpt-3 for few-shot knowledge-based vqa.
\newblock In \emph{Proceedings of the AAAI Conference on Artificial Intelligence}, pages 3081--3089, 2022.

\bibitem[Ye et~al.(2020)Ye, Zhan, Jiang, and Zhou]{ye2020heterogeneous}
Han-Jia Ye, De-Chuan Zhan, Yuan Jiang, and Zhi-Hua Zhou.
\newblock Heterogeneous few-shot model rectification with semantic mapping.
\newblock \emph{IEEE Transactions on Pattern Analysis and Machine Intelligence}, 43\penalty0 (11):\penalty0 3878--3891, 2020.

\bibitem[Ye et~al.(2021)Ye, Li, and Zhan]{ye2021task}
Han-Jia Ye, Xin-Chun Li, and De-Chuan Zhan.
\newblock Task cooperation for semi-supervised few-shot learning.
\newblock In \emph{Proceedings of the AAAI conference on artificial intelligence}, pages 10682--10690, 2021.

\bibitem[Ye et~al.(2022)Ye, Han, and Zhan]{ye2022revisiting}
Han-Jia Ye, Lu Han, and De-Chuan Zhan.
\newblock Revisiting unsupervised meta-learning via the characteristics of few-shot tasks.
\newblock \emph{IEEE Transactions on Pattern Analysis and Machine Intelligence}, 45\penalty0 (3):\penalty0 3721--3737, 2022.

\bibitem[Yuille(2019)]{YuilleLiuGradient2019DeepLearningVision}
Liu Yuille.
\newblock Limitations of deep learning for vision, and how we might fix them.
\newblock \emph{The Gradient}, 2019.

\bibitem[Zhang et~al.(2020)Zhang, Cai, Lin, and Shen]{Zhang_2020_CVPR}
Chi Zhang, Yujun Cai, Guosheng Lin, and Chunhua Shen.
\newblock Deepemd: Few-shot image classification with differentiable earth mover's distance and structured classifiers.
\newblock In \emph{IEEE/CVF Conference on Computer Vision and Pattern Recognition (CVPR)}, 2020.

\bibitem[Zhang et~al.(2021{\natexlab{a}})Zhang, Ding, Lin, Li, Wang, and Shen]{Zhang_2021_ICCV}
Chi Zhang, Henghui Ding, Guosheng Lin, Ruibo Li, Changhu Wang, and Chunhua Shen.
\newblock Meta navigator: Search for a good adaptation policy for few-shot learning.
\newblock In \emph{Proceedings of the IEEE/CVF International Conference on Computer Vision (ICCV)}, pages 9435--9444, 2021{\natexlab{a}}.

\bibitem[Zhang et~al.(2021{\natexlab{b}})Zhang, Song, Lin, Zheng, Pan, and Xu]{Zhang_2021_CVPR}
Chi Zhang, Nan Song, Guosheng Lin, Yun Zheng, Pan Pan, and Yinghui Xu.
\newblock Few-shot incremental learning with continually evolved classifiers.
\newblock In \emph{Proceedings of the IEEE/CVF Conference on Computer Vision and Pattern Recognition (CVPR)}, pages 12455--12464, 2021{\natexlab{b}}.

\bibitem[Zhang et~al.(2022)Zhang, Cai, Lin, and Shen]{zhang2022deepemd}
Chi Zhang, Yujun Cai, Guosheng Lin, and Chunhua Shen.
\newblock Deepemd: Differentiable earth mover's distance for few-shot learning.
\newblock \emph{IEEE Transactions on Pattern Analysis and Machine Intelligence}, 45\penalty0 (5):\penalty0 5632--5648, 2022.

\bibitem[Zhang et~al.(2023)Zhang, Huang, Jin, and Lu]{zhang2023vision}
Jingyi Zhang, Jiaxing Huang, Sheng Jin, and Shijian Lu.
\newblock Vision-language models for vision tasks: A survey.
\newblock \emph{arXiv preprint arXiv:2304.00685}, 2023.

\bibitem[Zhou et~al.(2022{\natexlab{a}})Zhou, Yang, Loy, and Liu]{zhou2022conditional}
Kaiyang Zhou, Jingkang Yang, Chen~Change Loy, and Ziwei Liu.
\newblock Conditional prompt learning for vision-language models.
\newblock In \emph{Proceedings of the IEEE/CVF Conference on Computer Vision and Pattern Recognition}, pages 16816--16825, 2022{\natexlab{a}}.

\bibitem[Zhou et~al.(2022{\natexlab{b}})Zhou, Yang, Loy, and Liu]{zhou2022learning}
Kaiyang Zhou, Jingkang Yang, Chen~Change Loy, and Ziwei Liu.
\newblock Learning to prompt for vision-language models.
\newblock \emph{International Journal of Computer Vision}, 130\penalty0 (9):\penalty0 2337--2348, 2022{\natexlab{b}}.

\end{thebibliography}
}

\clearpage
\setcounter{page}{1}

\maketitlesupplementary
\renewcommand{\thesection}{\Alph{section}}
\renewcommand{\thesubsection}{\thesection.\arabic{subsection}}

\section{Customized Description Prompts}

\begin{figure}[!h]
  \centering
    \includegraphics[width=\linewidth]{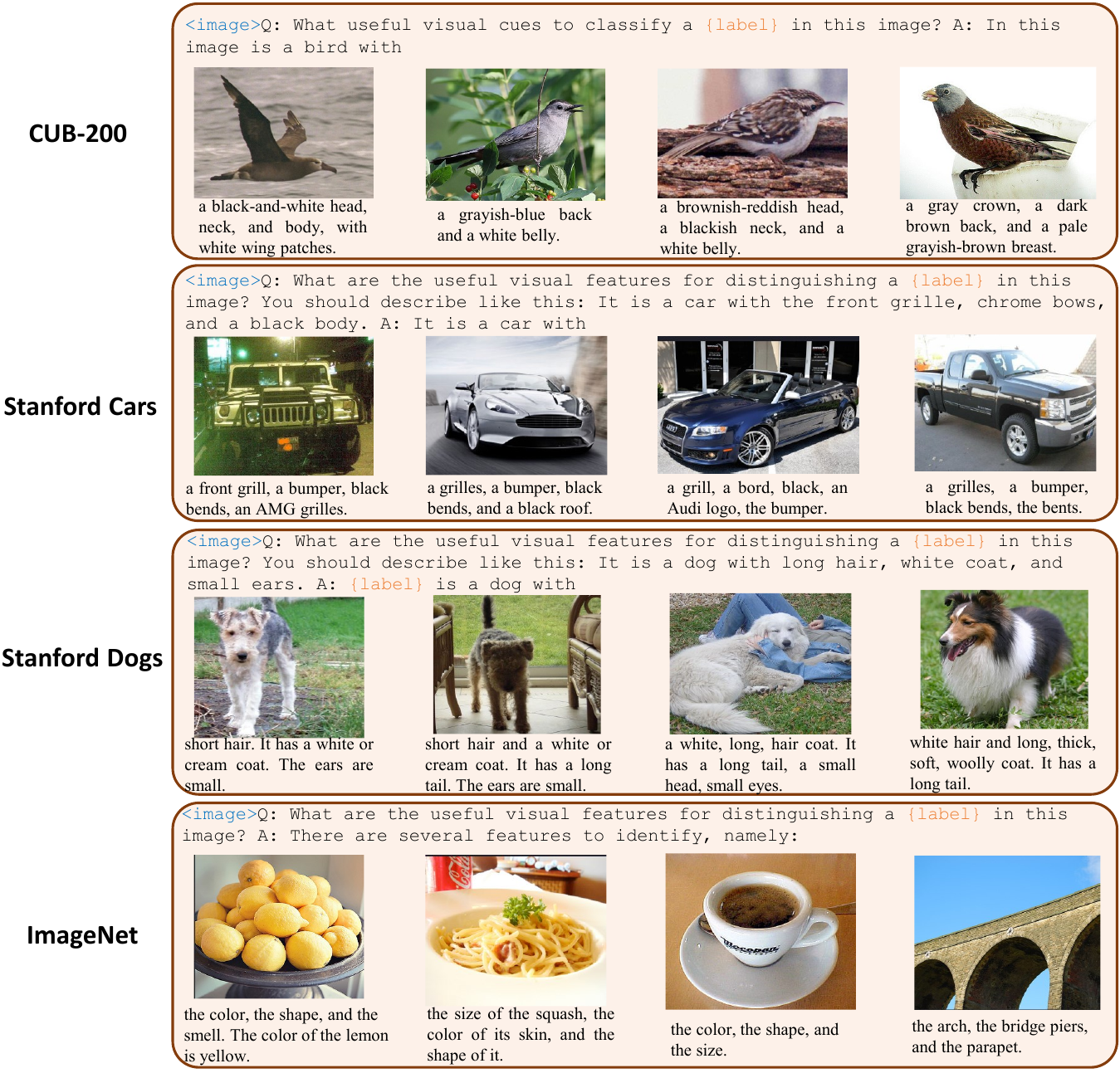}
    \caption{This is an example of prompt design tailored for each dataset. Notably, We incorporate example texts that exemplify the desired output format and content quality. }
    \label{fig:dd}
\end{figure}

Employing a generic prompt, such as "\textit{Q: What are the useful visual features for distinguishing a label in the image? A: There are several features to identify:}", may not consistently yield desirable descriptive texts for certain datasets. To rectify this,  we have crafted distinct prompts for each dataset, guiding the Vision-Language Models (VLMs) to produce more refined and discerning visual descriptions. These nuanced descriptions are integral to our Visual Description Enhancement (VDE) strategy. It subtly manipulates the label space through embedded visual cues, thereby enhancing the in-context classification performance. Fig.~\ref{fig:dd} displays examples of visual descriptions elicited by these custom-designed prompts for each specific dataset.

\section{Recognition of New Classes}
\begin{figure}[h]
  \centering
    \includegraphics[width=\linewidth]{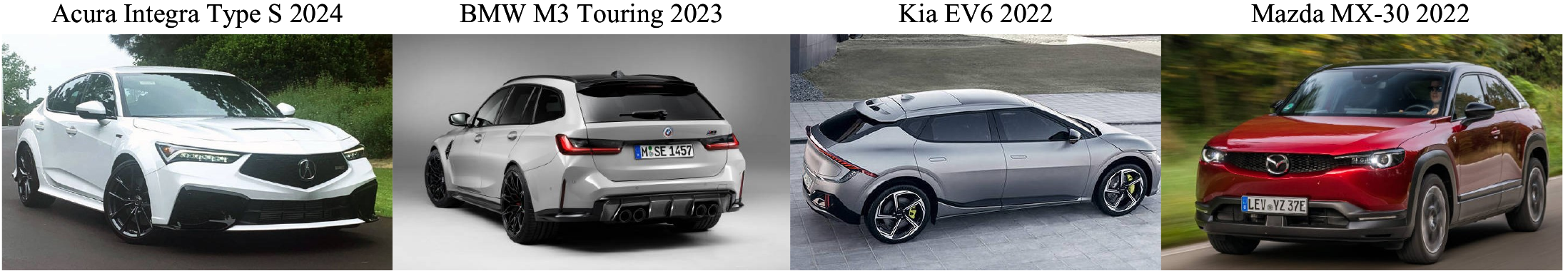}
    \caption{New class samples. }
    \label{fig:car}
\end{figure}

In our experiments, we explored the ability to learn with fewer samples against never-before-seen vehicle categories by collecting photos of new models produced or to be produced from 2022 to 2024 from the web, totaling 4 new categories (refer to Fig.~\ref{fig:car}) and 80 photos. For each new category, we selected 16 photos as a support set (for learning) and 4 photos as a test set (for evaluation). Furthermore, we conducted a comparative analysis of our proposed method's efficacy against the CLIP method, with the results detailed in Table~\ref{tab:supp}.

The results presented in Table~\ref{tab:supp} demonstrate the effectiveness of our method in categorizing never-before-seen categories, particularly in the challenging 1-shot and 2-shot learning settings. For instance, in 1-shot scenario, both the VDE method and the ENS. strategy achieve high scores of 15 or 16, showcasing their efficacy and robustness. This success can likely be attributed to the high-quality descriptions elicited by our meticulously designed prompts.
In contrast, the CLIP method, evaluated solely in the zero-shot learning setting, attains a score of 9. The performance of the SL strategy, on the other hand, fluctuated across settings, with a score of 10 in 1-shot setting and 16 in both 2-shot and 4-shot settings. the LDE strategy drops to a score of 8 in 2-shot learning, but then rises to a score of 15 in 4-shot scenario. These results underscore the fact that our approach in adapting to new categories, especially when data is limited, which is particularly important in domains such as image classification, where new categories emerge frequently. These findings suggest that certain strategies, especially VDE and ENS., are more effective in these scenarios.
\begin{table}[h]
\centering
\setlength{\extrarowheight}{0pt}
\addtolength{\extrarowheight}{\aboverulesep}
\addtolength{\extrarowheight}{\belowrulesep}
\setlength{\aboverulesep}{0pt}
\setlength{\belowrulesep}{0pt}

\begin{tabular}{ccccc} 
\toprule
\textbf{Strategy}                               & Zero-shot  & $1$-shot & $2$-shot & $4$-shot  \\ 
\midrule
CLIP                                            & $9$        & -        & -        & -         \\
SL                                              & -          & $10 $    & $16$     & $16$      \\
LDE(DD)                                             & -          & $9$      & $8$      & $15$      \\
\rowcolor[rgb]{0.937,0.937,0.937} \textbf{VDE}  & - & $\mathbf{15}$     & $\mathbf{16}$     & $\mathbf{16}$      \\
\rowcolor[rgb]{0.937,0.937,0.937} \textbf{ENS.} & - & $\mathbf{16}$     & $\mathbf{16}$     & $\mathbf{16}$      \\
\bottomrule
\end{tabular}
\caption{This table shows the number of novel samples correctly identified by different strategies in zero-shot, one-shot, two-shot, and four-shot learning settings.}
\label{tab:supp}
\end{table}

\end{document}